\documentclass[11pt,a4paper]{article}
\usepackage[hyperref]{naaclhlt2019}
\usepackage{times}
\aclfinalcopy
\usepackage{latexsym}
\usepackage{amsmath}
\usepackage{graphicx}
\usepackage{url} 
\usepackage{float}

\definecolor{mygreen}{rgb}{0.35, 0.60, 0.30}

\title{Jointly Optimizing Diversity and Relevance in Neural Response Generation}

\author{Xiang Gao \quad\quad Sungjin Lee \quad\quad Yizhe Zhang   \\ \textbf{Chris Brockett}\quad\quad \textbf{Michel Galley}\quad\quad \textbf{Jianfeng Gao}\quad\quad \textbf{Bill Dolan}\\
  Microsoft Research, Redmond, WA, USA \\
  {\small \tt \{xiag,sule,yizzhang,chrisbkt,mgalley,jfgao,billdol\}@microsoft.com}
}
\date{}

\begin{document}
\maketitle
\begin{abstract}
Although recent neural conversation models have shown great potential, they often generate bland and generic responses. While various approaches have been explored to diversify the output of the conversation model, the improvement often comes at the cost of decreased relevance \cite{zhang2018gan}. 
In this paper, we propose a \textsc{SpaceFusion} model
to jointly optimize diversity and relevance that essentially fuses the latent space of a sequence-to-sequence model and that of an autoencoder model by leveraging novel regularization terms. As a result, our approach induces a latent space in which the distance and direction from the predicted response vector roughly match the relevance and diversity, respectively. This property also lends itself well to an intuitive visualization of the latent space. Both automatic and human evaluation results demonstrate that the proposed approach brings significant improvement compared to strong baselines in both diversity and relevance. \footnote{An implementation of our model is available at \url{https://github.com/golsun/SpaceFusion}}

\end{abstract}

\section{Introduction}
\label{sec:intro}
\label{sec:intro}

The field of neural response generation is advancing rapidly both in terms of research and commercial applications \cite{gao2019neural, xiaoice, dstc7all, yizhe2019consistent}.  Nevertheless, vanilla sequence-to-sequence (S2S) models often generate  bland  and  generic responses \cite{li2016mmi}.
\begin{figure}[t]
    \centering
    \includegraphics[width=0.47\textwidth]{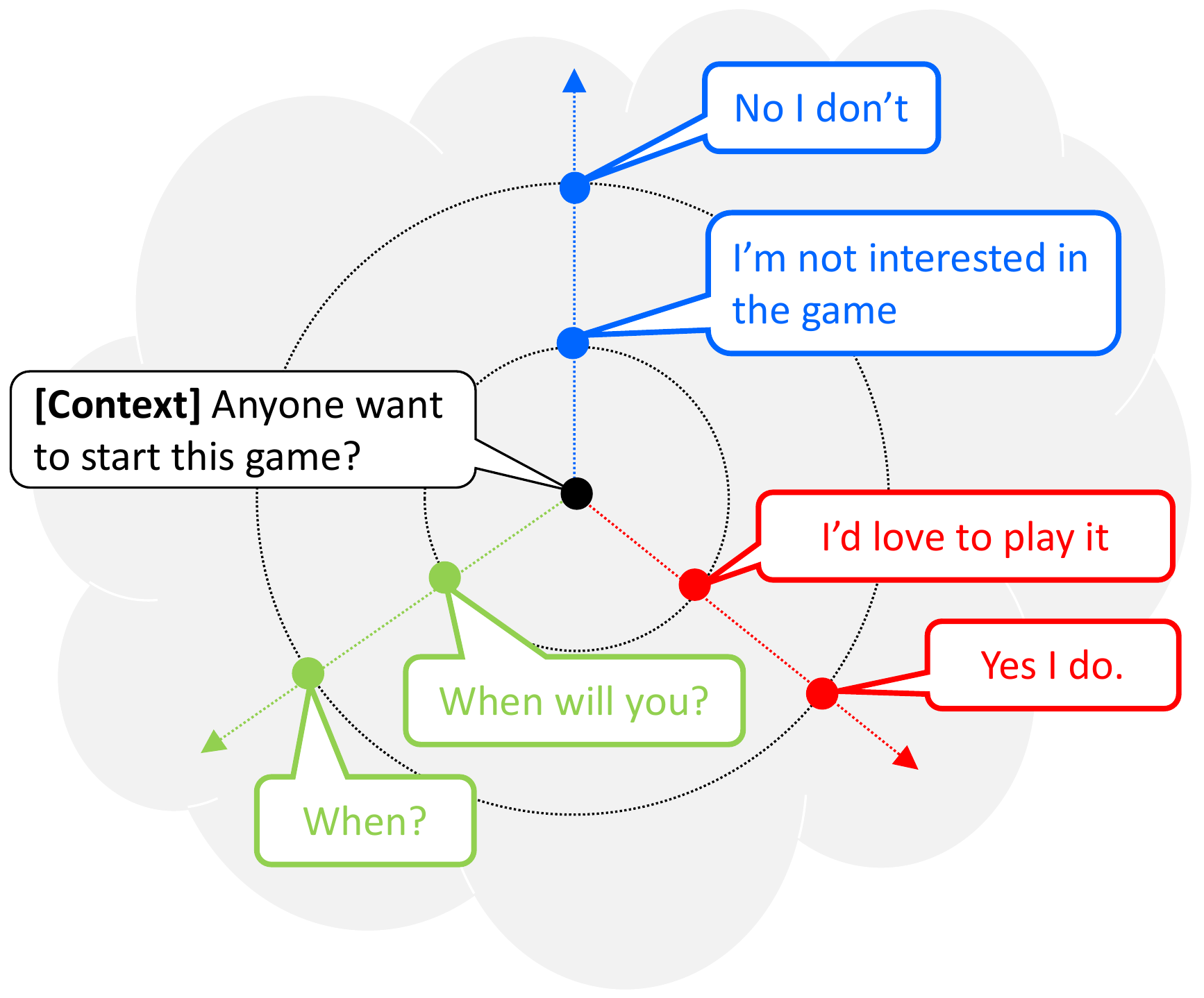}
    \caption{Illustration of one \textbf{context} and its multiple \textbf{ \textcolor{blue}{res}\textcolor{red}{pon}\textcolor{mygreen}{ses}} in the latent space induced by our model. Distance and direction from the predicted response vector given the context roughly match the relevance and diversity, respectively. Based on the example in Table~\ref{table:direction}.\footnote{}}
    \label{fig:intro}
\end{figure}
\footnotetext{For simplicity, we omitted the response at the center: "I would love to play this game". See Table~\ref{table:direction} for more details.}
\citet{li2016mmi} encourage diversity by re-ranking the beam search results according to their mutual information with the conversation context. However, as beam search itself often produces lists of nearly identical sequences, this method can require a large beam width (e.g. 200). As a result, re-ranking can be extremely time-consuming, raising difficulties for real-time applications. This highlights the need to improve the diversity 
of candidates before re-ranking, and the need to optimize for diversity during training rather than just at the decoding stage. 

While various approaches have been explored to diversify the output of  conversation models, the improvement often comes at the cost of decreased response relevance along other dimensions.
For instance, \citet{zhao2017cvae} present an approach to enhancing diversity by mapping diverse responses to a probability distribution
using a conditional variational autoencoder (CVAE). Despite the improved response diversity, this approach reduces response relevance as measured against the baseline.
One possible reason for this diversity-relevance trade-off is that such probabilistic approaches are not explicitly encouraged to induce a 
disentangled representation in latent space for controlling diversity and relevance independently. 
Consider a Gaussian distribution, which is widely used for CVAE. A Gaussian distribution naturally brings frequent responses near its mean and such responses are often generic and boring. 
To generate diverse and interesting responses, one needs to sample a little distance from the mean. But doing so naturally leads to infrequent and thus even irrelevant responses.

In this paper, we propose a novel geometrical approach that explicitly encourages a structured latent space in which the distance and direction from a predicted response vector roughly match the relevance and diversity, respectively, as illustrated in Figure~\ref{fig:intro}. 
To induce such a latent space, we leverage two different models: 1) a S2S model, producing the predicted response vector (the black dot at the center in Figure~\ref{fig:intro}), and 2) an autoencoder (AE) model, yielding the vectors for potential responses (the colored dots). In order to make the S2S and AE share the same latent space (the cloud), we use the same decoder for both and train them jointly end-to-end with novel regularization terms. As this fuses the two latent spaces, we refer to our model as \textsc{SpaceFusion}.

Regularization is necessary because only sharing the decoder, as in \cite{luan2017mtask}, does not necessarily align the latent spaces obtained by S2S and AE respectively or impose a disentangled structure onto the space. We introduce two regularization terms to tackle this issue. 1) interpolation term: we encourage a smooth semantic transition along the path between the predicted response vector and each target response vector (arrowed lines in Figure~\ref{fig:intro}). This term effectively prevents semantically 
different responses from aligning in the same direction, essentially scattering them over different directions. 2) fusion term: we want the vectors from the two models to be distributed in a homogeneous manner, rather than forming two separate clusters (Figure~\ref{fig:vis_fusion}) that can potentially make sampling non-trivial. 
With the resulting latent space, we can control relevance and diversity by respectively adjusting distance and direction from a predicted response vector, without sacrificing each other greatly.

Our approach also lends itself well to the intuitive visualization of latent space.
Since our model allows us to geometrically find not only the predicted response vector but also the target response vector as in Figure~\ref{fig:vis_fusion}, we can visually interpret the structure of latent space and identify major issues thereof. We devote Section~\ref{sec:vis} to show comprehensive examples for visualization-based analysis.

Automatic and human evaluations demonstrate that the proposed approach improves both the diversity and relevance of the responses, compared to strong baselines on two datasets with one-to-many context-response mapping.

\section{Related Work}

\paragraph{Grounded conversation models} utilize extra context inputs besides conversation history, such as persona \cite{li2016persona}, textual knowledge \cite{ghazvininejad2017knowledge, dstc7track2}, dialog act \cite{zhao2017cvae} and emotion \cite{huber2018emotional}. Our approach does not depend on such extra input and thus is complementary to this line of studies.

\paragraph{Variational autoencoder (VAE) models} explicitly model the uncertainty of responses in latent space. \citet{bowman2016vae} used VAE with Long-Short Term Memory (LSTM) cells to generate sentences. The basic idea of VAE is to encode the input $x$ into a probability distribution (e.g. Gaussian) $z$ instead of a point encoding. 
However, it suffers from the \textit{vanishing latent variable problem} \cite{bowman2016vae, zhao2017cvae} when applied to text generation tasks. \citet{bowman2016vae, fu2019cyclical} proposed to tackle this problem with word dropping and specific KL annealing methods. 
\citet{zhao2017cvae} proposed to add a \textit{bag-of-word} loss, complementary to KL annealing. Applying this to a CVAE conversation model, they showed that even greedy decoding can generate diverse responses. However, as VAE/CVAE conversation models can be limited to a simple latent representations such as standard Gaussian distribution, \citet{gu2018dialogwae} proposed to enrich the latent space by leveraging a Gaussian mixture prior. Our work takes a geometrical approach that is fundamentally different from probabilistic approaches to tackle the limitations of parameteric distributions in representation and difficulties in training.

\paragraph{Decoding and ranking} encourage diversity during the decoding stage. As ``vanilla" beam search often produces lists of nearly identical sequences, \citet{vijayakumar2016dbs} propose to include a dissimilarity term in the objective of beam search decoding. \citet{li2016mmi} re-ranked the results obtained by beam search based on mutual information with the context using a separately trained response-to-context S2S model.

\paragraph{Multi-task learning} is another line of studies related to the present work (see Section~\ref{sec:model}). \citet{sennrich2016nmt_mtask} use multi-task learning to improve neural machine translation by utilizing monolingual data, which usually far exceeds the amount of parallel data. A similar idea is applied by \citet{luan2017mtask} to conversational modeling, involving two tasks: 1) a S2S model that learns a context-to-response mapping using conversation data, and 2) an AE model that utilizes speaker-specific non-conversational data. The decoders of S2S and AE were shared, and the two tasks were trained alternately.

\section{The \textsc{SpaceFusion} Model}
\subsection{Problem statement}
Let $\mathcal{D}=[(x_0,y_0),(x_1,y_1),\cdots,(x_n,y_n)]$ denote a conversational dataset, where $x_i$ and $y_i$ are a context and its response, respectively. $x_i$ consists of one or more utterances. Our aim is to train a model on $\mathcal{D}$ to generate relevant and diverse responses given a context.

\subsection{Fusing latent spaces}
\label{sec:model}

We design our model to induce a latent space where different responses for a given context are in different directions around the predicted response vector, as illustrated in Figure~\ref{fig:intro}. Then we can obtain diverse responses by varying the direction and keep their relevance by sampling near the predicted response vector. 

\begin{figure}[tb]
    \centering
    \includegraphics[width=0.47\textwidth]{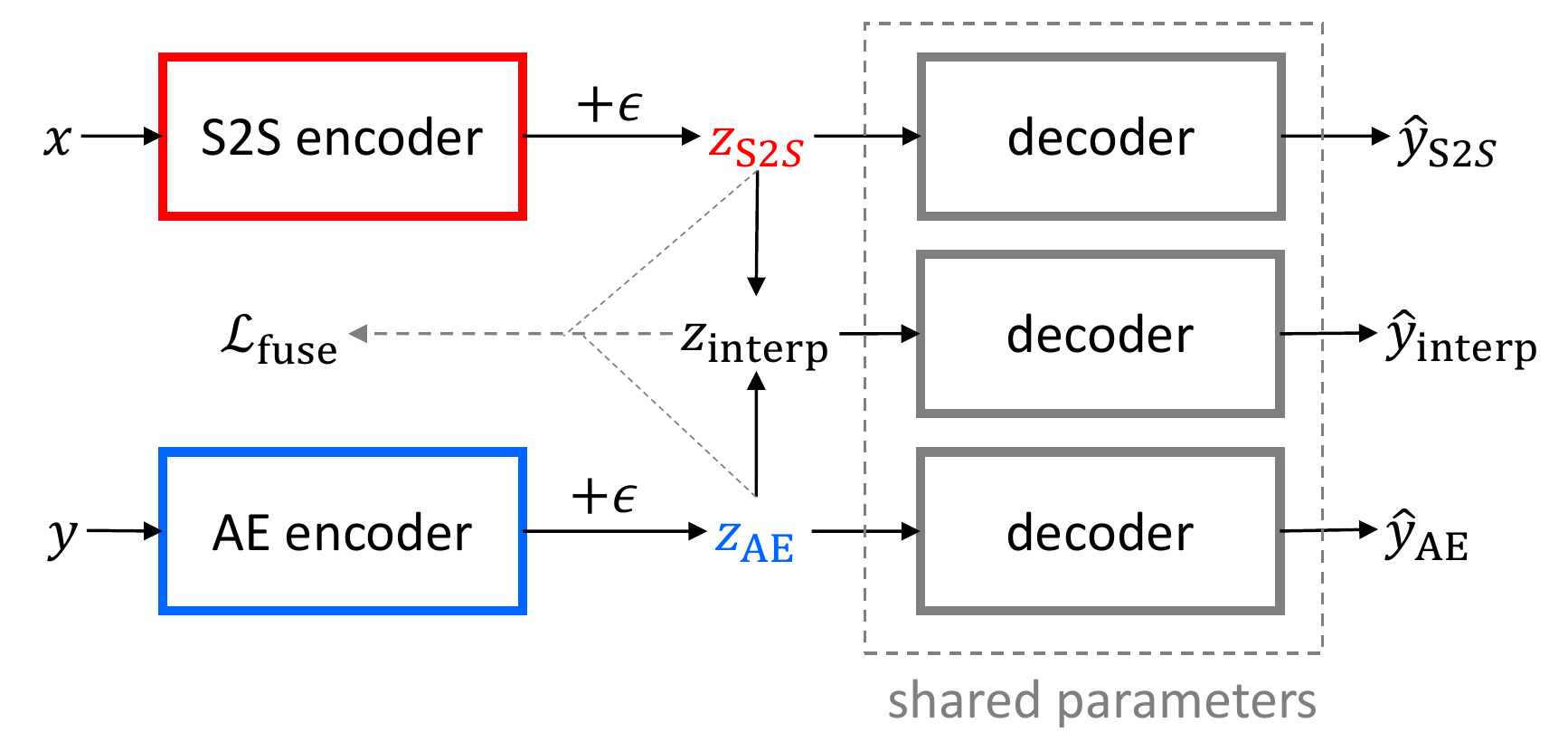}
    \caption{\textsc{SpaceFusion} model architecture.}
    \label{fig:model_train}
\end{figure}

To fulfill this goal, we first produce the predicted response representation $z_{\text{S2S}}$ and target response representations $z_{\text{AE}}$ using an S2S model and an AE model, respectively, as illustrated in Figure~\ref{fig:model_train}. Both encoders are implemented using stacked Gated Recurrent Unit (GRU) \cite{cho2014gru} cells followed by a noise layer that adds multivariate Gaussian noise $\epsilon \sim N(0,\sigma ^ 2 \mathbf{I})$. 
We then explicitly encourage smooth semantic transition along the path from $z_{\text{S2S}}$ to $z_{\text{AE}}$ by imposing any interpolation between them to generate the same response via the following loss term:
\begin{align} 
    \mathcal{L}_{\text{interp}} & = - \dfrac{1}{|y|} \log p(y|z_{\text{interp}}) 
\end{align}
where $z_{\text{interp}} = u z_{\text{S2S}} + (1-u)z_{\text{AE}}$ and $u \sim U(0,1)$ is a uniformly distributed random variable. $|y|$ is the number of words in $y$. Note that it is this regularization term that effectively prevents significantly different responses from aligning in the same direction, essentially scattering them over different directions.
In order for this interpolation loss to work, we share the same decoder for both AE and S2S models as in \cite{luan2017mtask}. The decoder consists of stacked GRU cells followed by a softmax layer. 
It is worth mentioning that $z_{\text{interp}}$ is not just randomly drawn from a single line but from a richer probabilistic region as both $z_{\text{interp}}$ and $z_{\text{S2S}}$ are stochastic due to the random component $\epsilon$.

Now, we want vectors from both the AE and S2S models to be distributed in a homogeneous manner scattered over the entire space while keeping the distance between $z_{\text{S2S}}$ and $z_{\text{AE}}$ as small as possible for any (context-response) pair in the training data. This objective is represented in the following regularization term:
\begin{align}
     \mathcal{L}_{\text{fuse}} & = \sum_{i\in \text{batch}} \frac{d( z_{\text{S2S}}(x_i),z_{\text{AE}}(y_i) )}{n} \nonumber \\
        & - \sum_{i,j\in \text{batch}, i \neq j} \frac{d(z_{\text{S2S}}(x_i), z_{\text{S2S}}(x_j))}{n^2-n} \nonumber \\
        & - \sum_{i,j\in \text{batch}, i \neq j} \frac{d(z_{\text{AE}}(y_i), z_{\text{AE}}(y_j))}{n^2-n}
        \label{eq:L_fuse}
\end{align}
where $n$ is the batch size and $d(a,b)$ is the root mean square of the difference between $a$ and $b$. For each batch, we basically disperse vectors obtained by the same model and pull the predicted response vectors to the corresponding target response vectors.
In practice, we found that the performance is better if the Euclidean distance is clipped to a prescribed maximum value.\footnote{This value is set as 0.3 for the present experiments}

Finally, with weight parameters $\alpha$ and $\beta$, the loss function is defined as: 
\begin{align}  
\label{eq:final_loss}  
    \mathcal{L} =   & - \frac{1}{|y|} \log p(y|z_{\text{S2S}}) \nonumber \\ 
                    & - \frac{1}{|y|} \log p(y|z_{\text{AE}}) \nonumber\\ 
                    & + \alpha \mathcal{L}_{\text{interp}} + \beta \mathcal{L}_{\text{fuse}}
\end{align}
As $\mathcal{L}_{\text{interp}}$ and $\mathcal{L}_{\text{fuse}}$ encourage the path between $z_{\text{S2S}}$ and $z_{\text{AE}}$ to be smooth and short while scattering vectors over the entire space, they effectively fuse the $z_{\text{S2S}}$ latent space and the $z_{\text{AE}}$ latent space. Accordingly we refer this approach as \textsc{SpaceFusion} with path regularization.

\subsection{Training}
In contrast to previous multi-task conversation model \cite{luan2017mtask}, where S2S and AE are trained alternately, our approach trains S2S and AE at the same time by minimizing the loss function of Equation \ref{eq:final_loss}.

\subsection{Inference}
Like \citet{zhao2017cvae, bowman2016vae}, for a given context, we sample different latent vectors to obtain multiple hypotheses. This is done by adding a random vector $r$ that is uniformly sampled from a hypersphere of radius $|r|$ to the prediction $z_{\text{S2S}}(x)$.
\begin{align} 
\label{eq:infer}
z(x,r)=z_{\text{S2S}}(x)+r
\end{align}
\label{sec:infer}
 where $|r|$ is tuned on the validation set to optimize the trade-off between relevance and diversity. $z(x,r)$ is then fed to the decoder as the initial state of GRU cells. We then generate responses using greedy decoding.\footnote{Although we use greedy decoding in this work, other decoding techniques, such as beam search, can be applied.}

\section{Experiment Setup}
\subsection{Datasets}
We used the following datasets. Some of their key features are presented in Table~\ref{table:dataset}.

\paragraph{Switchboard:} We use the version offered by \citet{zhao2017cvae}, which is an extension of the original version by \citet{godfrey1997switchboard}. \citet{zhao2017cvae} collected multiple references for the test set using information retrieval (IR) techniques followed by human filtering, and randomly
split the data into 2316/60/62 conversations for
train/validate/test, respectively. Each conversation has multiple turns and thus multiple $(x,y)$ pairs, as listed in Table~\ref{table:dataset}. As our approach does not utilize extra information except conversation history, we removed the meta data (e.g. gender, age, prompt) from this dataset.

\paragraph{Reddit:} As the Switchboard dataset is relatively small and multiple references are synthetically constructed, we have developed another multi-reference dataset by extracting posts and comments on Reddit.com during 2011 collected by a third party.\footnote{\url{http://files.pushshift.io/reddit/comments/}} As each Reddit post and comment may have multiple comments, it is a natural source of multi-reference responses. We further filtered the data based on the number of replies to obtain the final conversation dataset in which each context has at least 10 different responses, and on average the number of responses is 24.1 for a given context. The size is significantly larger than Switchboard, as listed in Table~\ref{table:dataset}. The conversations are randomly shuffled before being split into train/valid/test subsets.

\begin{table}[t]
    \centering
    \small
    \begin{tabular}{p{0.16\textwidth}|p{0.12\textwidth}p{0.12\textwidth}} \hline 
    
                 &  Switchboard    & Reddit \\ \hline
    train ($x,y$) samples & 0.2M          & 7.3M \\
    test ($x,y$) samples & 5418 & 5000\\ \hline
    ref. source & IR+filtering  & natural \\
    ref. availability & test only  & train/vali/test \\
    ref. per context  & 7.7 & 24.1\\ \hline
    \end{tabular}
    \caption{Key features of the datasets.}
    \label{table:dataset}
\end{table}
\subsection{Model setup}
Both encoders and the shared decoder consist of two GRU cells, each with 128 hidden units. The variance of the noise layer in each decoder is $\sigma^2=0.1^2$.
The word embedding dimension is 128. The weight parameters 
(see Equation~\ref{eq:final_loss}) are set as $\alpha=1$ and $\beta=30$. For both datasets, the inference radius $|r|$ (see Equation~\ref{eq:infer}) is set to 1.5 which optimizes F1 score on the validation set.
All models are trained using the Adam method \cite{kingma2014adam} with a learning rate of 0.001 on both datasets until convergence (around 4 epochs for Reddit and 10 epochs for Switchboard).

\subsection{Automatic evaluation}
For a given context $x$, we have $N_r$ reference responses and generate the same number of hypotheses.\footnote{We set the number of hypotheses equal to the number of references to encourage precision and recall have comparable impact on F1} We define the following metrics based on 4-gram BLEU \cite{papineni2002bleu}, as suggested by \citet{zhao2017cvae}.
\begin{align*}
    \text{Precision} &= \frac{1}{{N_r}}\sum_{i=1}^{N_r} \max_{j\in[1,N_r]}\text{BLEU}(r_j, h_i)  \\
    \text{Recall} &= \frac{1}{{N_r}}\sum_{j=1}^{N_r} \max_{i\in[1,N_r]}\text{BLEU}(r_j,h_i) \\
    \text{F1} &= 2\,\frac{ \text{precision} \cdot \text{recall} } {\text{precision} +  \text{recall}} \\
\end{align*}

\begin{table*}[ht]
    \centering
    \small
    \begin{tabular}{p{0.03\textwidth} p{0.25\textwidth}|p{0.03\textwidth} p{0.25\textwidth}|p{0.03\textwidth} p{0.25\textwidth}}  
    
    \hline
    
    \multicolumn{6}{c}{context $x$: Anyone want to start this game?}\\
    \multicolumn{6}{c}{response at $u=0$: I would love to play this game.}\\
    
    \hline
    $u$   & \textbf{towards ``No I don't."}           & $u$    & \textbf{towards ``when?"}                 & $u$    & \textbf{towards ``Yes I do."} \\ \hline
    0.18  & I am not interested in the game. & 0.15   & I'd be interested in the game    & 0.15   & I'd love to play it. \\
    0.21  & I am not interested.             & 0.31   & When is it?                      & 0.27   & Yes I do. \\
    0.30  & No I don't.                      & 0.40   & When will you?                   &        & \\
          &                                  & 1.00   & When?                            &        & \\ \hline
    
    \end{tabular}
    \caption{Semantic interpolation along different directions $y$. Results decoded from $z_{\text{interp}}$ 
    See Fig.~\ref{fig:intro} for a visualization.}.
    \label{table:direction}
\end{table*}

We use Precision as an approximate surrogate metric for relevance and Recall for diversity. It should be noted that recall is not equivalent to other diversity metrics, e.g., distinct \cite{li2016mmi} and entropy \cite{zhang2018gan}, which only depend on hypotheses. One potential issue of these metrics is that even randomly generated responses may yield a high diversity score. F1 is the harmonic average of these two and is used to measure the overall response quality.

\subsection{Human evaluation}

We conduct a human evaluation using crowdworkers. For each hypothesis, given its context, we ask three annotators to individually measure the quality, on a scale of 1 to 5, in terms of two aspects: relevance and interest.
Interestingness is treated as an estimation of the diversity, as these two are often correlated. The hypotheses from all systems are shuffled before being provided to annotators. System names are invisible to the annotators.

\subsection{Baselines}
We compare the proposed model with the following baseline models:
\paragraph{S2S+Sampling:} We consider a vanilla version of S2S model. The dimensions are similar to our model: both encoder and decoder consist of two stacked GRU cells with 128 hidden units, and the word embedding size is 128. As in the baseline in \citet{zhao2017cvae}, we applied softmax sampling at inference time to generate multiple hypotheses.

\paragraph{CVAE+BOW:} For the CVAE conversation model, we use the original implementation and hyperparameters of \citet{zhao2017cvae} with the bag-of-words (BOW) loss. The number of trainable model parameters is 15.4M, which is much larger than our model (3.2M).

\paragraph{MTask:} Since our approach utilizes a multi-task learning scheme, we also compare it against a vanilla multi-task learning model, MTask, similar to \cite{luan2017mtask}, to illustrate the effect of space fusion. The model architecture and hyperparameters are identical to the proposed model except that the loss function is $\mathcal{L} = -\log p(y|z_{\text{S2S}}) -\log p(y|z_{\text{AE}})$.

\section{Results and Analysis}
\begin{table*}[!ht]
    \centering
    \small
    \begin{tabular}{p{0.03\textwidth} p{0.35\textwidth}|p{0.03\textwidth} p{0.35\textwidth}}  
    
    \hline
    
    \multicolumn{4}{c}{context $x$: Anyone want to start this game?}\\
    
    \multicolumn{4}{c}{towards one possible target $y$: Yes I do.} \\ 
    \hline
    $u$   & \textbf{with regularization}             & $u$    & \textbf{without regularization} \\ \hline
    0.00  & I would love to play this game.  & 0.00   & I would have to play with the game. \\
    0.15  & I would love to play it.         & 0.29   & Dude, I know, but, or etc.  \\
    0.30  & Yes I do                         & 0.61   & Op I was after though today\\
          &                                  & 0.85   & I'm single :( though \\
          &                                  & 0.90   & Yes I do. \\
    \hline

    \end{tabular}
    \caption{Semantic interpolation with and without regularization. Results decoded from $z_{\text{interp}}$ 
    .}
    \label{table:interp}
\end{table*}

\subsection{In-depth analysis of latent space}

\label{sec:vis}
In this section, we undertake an in-depth analysis to verify whether the latent space induced by our method manifests desirable properties, namely: 1) disentangled space structure between relevance and diversity, 2) homogeneous space distribution in which semantics changes smoothly without holes.
We first provide a qualitative investigation based on real examples. Then, we present a set of corpus-level quantitative analyses focused on  geometric properties.

\subsubsection{Qualitative examples} 
In Table~\ref{table:direction}, we investigate three different directions from the context ``\texttt{Anyone want to start this game?}''
, which is a real example taken from Reddit.
The three different directions correspond to clearly different semantics: ``\texttt{No I don't}", ``\texttt{when?}" and ``\texttt{Yes I do.}" If we generate a response with the vector predicted by the S2S model ($u=0$), our model outputs ``\texttt{I would love to play this game}" which is highly relevant to the context. Now as we move along each direction, we can see our model gradually transforms the response toward the corresponding responses of each direction. For instance, towards ``\texttt{No I don't}", our model gradually transforms the response to ``\texttt{I am not interested in the game}" ($u=0.18$) and then ``\texttt{I am not interested.}" ($u=0.21$). In contrary, towards ``\texttt{Yes I do}", the response transforms to ``\texttt{I would love to play it.}" ($u=0.15$). Besides the positive or negative directions, the same transition applies to other directions such as ``\texttt{When?}". This example clearly shows that there is a rough correspondence between geometric properties and semantic properties in the latent space induced by our method as shown in Figure~\ref{fig:intro}-- the relevance of the response decreases as we move away from the predicted response vector and different directions are associated with semantically
different responses.

\subsubsection{Direction vs. diversity} 
\begin{figure}[ht]
    \centering
    \includegraphics[width=0.5\textwidth]{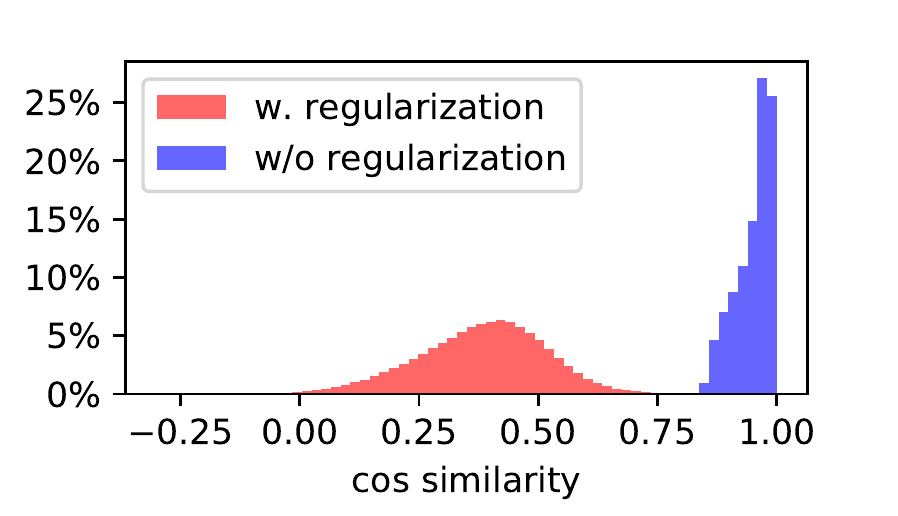}
    \caption{Distribution of the directions from a given context to its multiple responses, measured by the cosine similarity between $z_{\text{AE}}(y_{k,i})-z_{\text{S2S}}(x_k)$ and $z_{\text{AE}}(y_{k,j})-z_{\text{S2S}}(x_k)$. Histogram calculated based on 1000 $x_k$ from Reddit test data and visualized with bin width of 0.02.}
    \label{fig:vis_angel}
\end{figure}

In order to quantitatively verify the correspondence between \textit{direction} and \textit{diversity}, we visualize the distribution of cosine similarities among multiple references for each context for a set of 1000 random samples drawn from the test dataset. Specifically, for a context $x_k$ and its associated reference responses $[y_{k,0}, y_{k,1}, \cdots]$, we compute the cosine similarity between $z_{\text{AE}}(y_{k,i})-z_{\text{S2S}}(x_k)$ and $z_{\text{AE}}(y_{k,j})-z_{\text{S2S}}(x_k)$.
In Figure~\ref{fig:vis_angel}, we compare the distribution of our model with that of MTask, which does not employ our regularization terms.
While our method yields a bell shaped curve with average cosine similarity being close to zero (0.38), the distribution of MTask is extremely skewed with average cosine similarity being close to 1 (0.95). This indicates that the directions of the reference responses are more evenly distributed in our latent space whereas everything is packed in a narrow band in the MTask's space. This essentially makes the inference process simple and robust in that one can choose arbitrary directions to generate diverse responses.

\subsubsection{Distance vs. relevance} 
\begin{figure}[ht]
    \centering
    \includegraphics[width=0.45\textwidth]{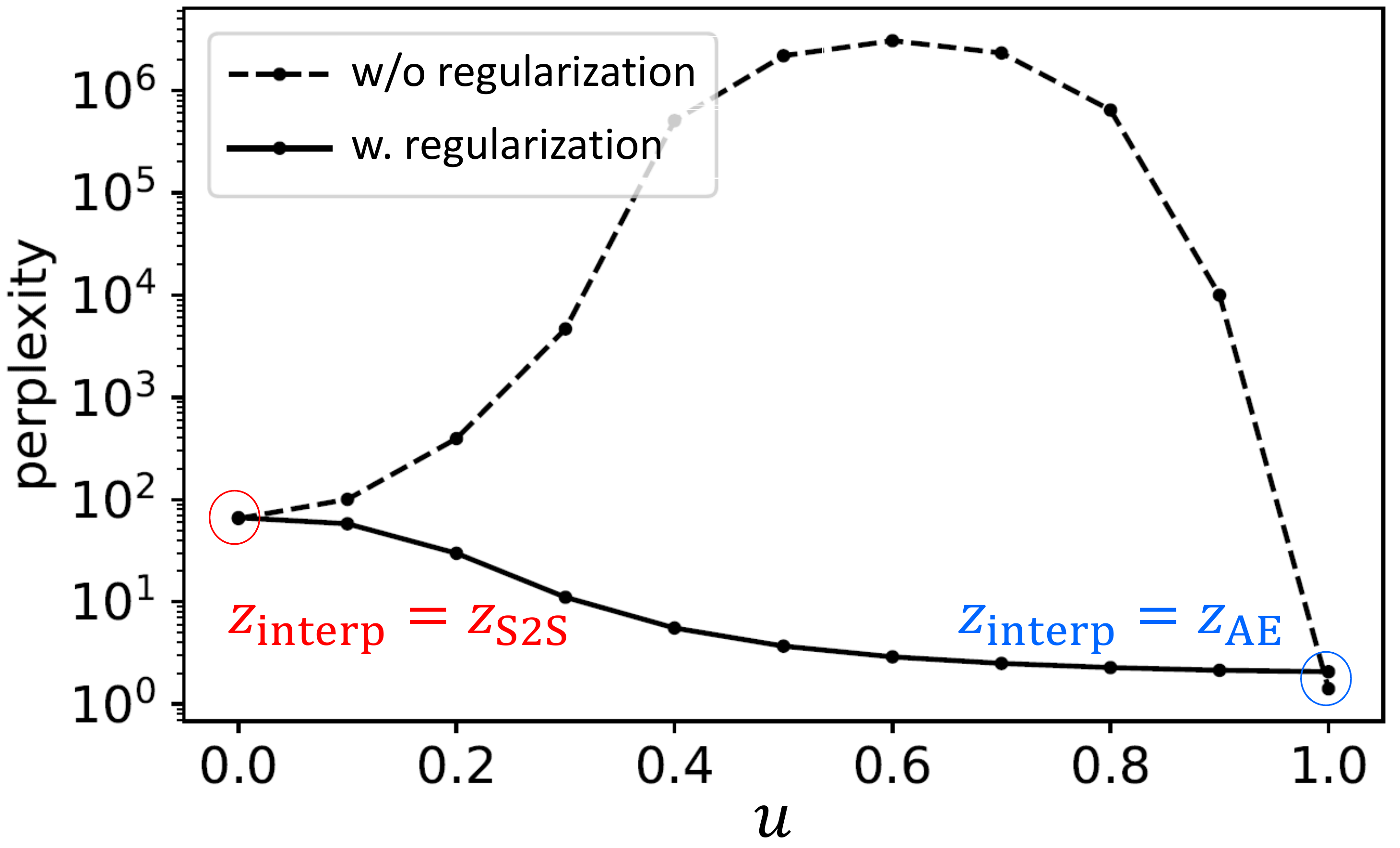}
    \caption{Perplexity of $z_{\text{interp}}$ on the Reddit test dataset as a function of $u$ for simple multi-task model (without regularization, dashed line) and \textsc{SpaceFusion} (with regularization, solid line).}
    \label{fig:vis_interp}
\end{figure}

In order to quantitatively verify the correspondence between \textit{distance} and \textit{relevance}, we visualize the perplexity of reference responses along the path from the associated $z_{\text{S2S}}$ ($u=0$) to the $z_{\text{AE}}$  ($u=1$) corresponding to the predicted response. In Figure~\ref{fig:vis_interp}, we compare our model with MTask, which as already noted, does not employ our regularization terms. While our model shows a gradual increase in perplexity, there is a huge bump for MTask's line. This clearly indicates that there is a rough correspondence between distance and relevance in our latent space whereas even a slight change can lead to an irrelevant response in the MTask's space.

We further illustrate the smooth change in relevance according to distance for a specific example in Table~\ref{table:interp}. Given the context ``\texttt{Anyone want to start this game?}", our model is able to transition from the predicted response ``\texttt{I would love to play this game}" to a one of reference responses ``\texttt{Yes I do}". The relevance smoothly descreases, generating intermediate responses such as ``\texttt{I would love to play it.}" In contrary, the MTask model tends to produce irrelevant or ungrammatical responses as it moves away from the predicted response.

\subsubsection{Homogeneity and Convexity} 
\begin{figure}[ht]
    \centering
    \includegraphics[width=0.5\textwidth]{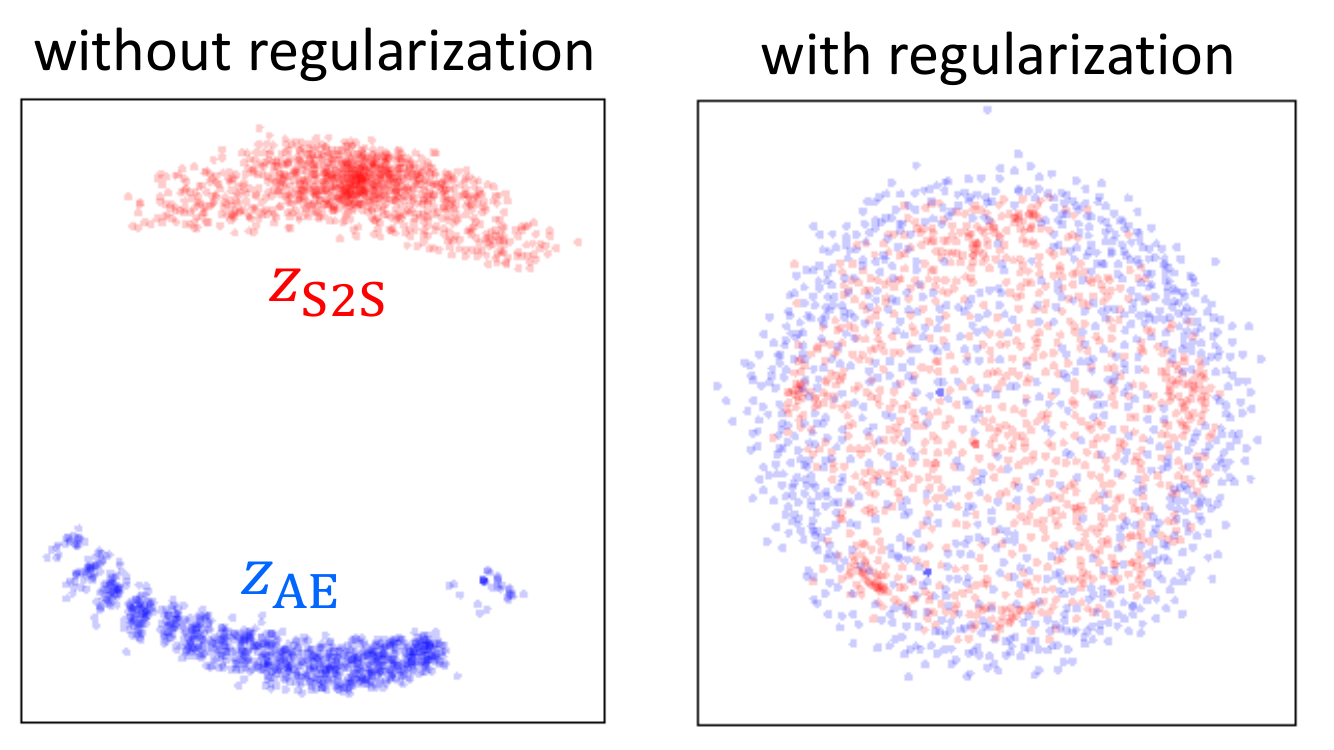}
    \caption{MDS visualization of the two latent spaces: $z_{\text{s2s}}$ (\textcolor{red}{red} dots) and $z_{\text{AE}}$ (\textcolor{blue}{blue} dots) of 1000 randomly picked ($x, y$) pairs from the Reddit test dataset. Left: multi-task model (without regularization); right: \textsc{SpaceFusion} (with regularization).}
    \label{fig:vis_fusion}
\end{figure}

Other desirable properties, with which we want to equip our latent space are \textit{homogeneity} and \textit{convexity}. If the space is not homogeneous, we have to sample differently depending on the regional traits. If the space is not convex, we have to worry about running into the holes that are not properly associated with valid semantic meanings.
In order to verify homogeneity and convexity, we visualize our latent space in a 2D space produced by the multidimensional scaling (MDS) algorithm~\cite{borg2003mds}, which approximately preserves pairwise distance. For comparison, we also provide a visualization for MTask. As shown in Figure~\ref{fig:vis_fusion}, our latent space offers great homogeneity and convexity regardless of which model is used to produce a dot (i.e. $z_{S2S}$ or $z_{AE}$). In contrary, MTask's latent space forms two separate clusters for $z_{S2S}$ and $z_{AE}$ with a large gap in-between where no training samples were mapped to.

\subsection{Automatic evaluation}

We let each system generate 100 hypotheses $\{h_j\}$ for each context $x_i$ in the test dataset. Assuming $x_i$ has $N_{r,i}$ references, we pick the top $N_{r,i}$ distinct hypotheses ranked by $\log p(h_j|x_i) + \lambda |h_j|$. Similar to \cite{li2016mmi, wu2016google}, we takes $|h_j|$ into consideration, as BLEU is sensitive to length. For fair comparison, $\lambda$ is tuned such that the average hypothesis length becomes roughly the same for all systems and approaches the average length of the references.\footnote{Approximately 10 words/tokens for Switchboard and 12 for Reddit}

The automatic evaluation results are reported in Table~\ref{table:auto}. On both datasets, the proposed system consistently outperforms the baselines by a large margin in Precision, Recall, and F1.

Examples of  system outputs and human references can be found in Table~\ref{table:example_reddit} and Table~\ref{table:example_switchboard} for Reddit and Switchboard, respectively. As shown in the examples, CVAE+BOW and other baseline models may generate diverse but not-so-relevant responses. 

\begin{table}[ht]
    \centering
    \small
    \begin{tabular}{p{0.1\textwidth}|p{0.12\textwidth}|p{0.06\textwidth}p{0.04\textwidth}p{0.03\textwidth}} \hline 
     
    dataset & model           &  Precision    & Recall        & F1 \\ \hline
                & \textsc{SpaceFusion}      & \textbf{1.22} & \textbf{0.66} &  \textbf{0.86} \\ 
    Switchboard & CVAE+BOW     & 0.76          & 0.57          & 0.65 \\
                & MTask   & 0.75          & 0.43          & 0.54 \\ 
                & S2S+Sampling & 0.57          & 0.48          & 0.52 \\ \hline
                & \textsc{SpaceFusion}      & \textbf{0.40} & \textbf{0.26} & \textbf{0.31} \\ 
    Reddit      & CVAE+BOW     & 0.16          & 0.18          & 0.17 \\
                & MTask   & 0.31          & 0.18          & 0.23 \\ 
                & S2S+Sampling & 0.10          & 0.11          & 0.11 \\ \hline
    
    \end{tabular}
    \caption{Performance of each model on automatic measures. The highest score in each row is in \textbf{bold} for each dataset. Note that our BLEU scores are normalized to $[0,100]$.}
    \label{table:auto}
\end{table}

\begin{table}[h]
    \centering
    \small
    \begin{tabular}{p{0.1\textwidth}|p{0.33\textwidth}} \hline 
    \textbf{context} & Everything about this movie is awesome!\\
    \hline
    \hline
    \textsc{Space}  & $\circ$ I love this movie.\\
    \textsc{Fusion} & $\circ$ It's so awesome!!! I have no idea how to watch this movie. I can't wait for the trailer.\\
        & $\circ$ I don't think i'm a fan of the movie.\\ 
        & $\circ$ I would love to see this.\\
        & $\circ$ I want to watch this movie.\\
    \hline
    CVAE & $\circ$  Smartphones of the best games!.\\
    +BOW    & $\circ$ I'm in the same boat! I feel the same way about this\\
        & $\circ$ I don't know why but can't tell if the New York will be the only one. \\
        & $\circ$ Caves would never say yes, but I'd love to know.\\
        & $\circ$ I know where this movie is. \\
    \hline
    MTask & $\circ$ This is so funny. \\
        & $\circ$ I think I saw this.  \\
        & $\circ$ I don't know how many people do you. \\
        & $\circ$ I would. \\
        & $\circ$ I remember watching was was disappointed. \\
    \hline
    S2S+ & $\circ$  Whoa man, this is amazing!!! \\
    sampling &  $\circ$ Man, I love Tom Cruise and I love the show. \\
        & $\circ$ Is this a bill on the right? \\
        & $\circ$ More like samurai jack. \\
        & $\circ$ I weep for the hivemind. \\
    \hline
    human & $\circ$ I love this movie.\\
        & $\circ$ Watched this yesterday! so amazing. \\
        & $\circ$ My favorite Muppet movie by far. \\
        & $\circ$ Seriously. Good times. Great oldies. \\
        & $\circ$ Now you have got the song into my head. \\
    \hline
    \end{tabular}
    \caption{Example system outputs and human reference on Reddit dataset}
    \label{table:example_reddit}
\end{table}

\begin{table}[h]
    \centering
    \small
    \begin{tabular}{p{0.07\textwidth}|p{0.37\textwidth}} \hline 
    \textbf{context} & \textbf{A}: Are they doing a lot of recycling out in ***? \\
            & \textbf{B}: Well at my workplace they are. We have places for aluminum cans and we have everybody been issued a separate trash can for recyclable paper. Let's see them get that word straight. \\
            & \textbf{A}: Uh-huh \\
            & \textbf{B}: As far as the community goes, it's pretty much voluntary. Nothing's been done. ***. We have just private garbage services. \\
    \hline
    \hline
    \textsc{Space} & $\circ$ They're not recycling as a matter of fact.\\
    \textsc{Fusion}    & $\circ$ And it's a waste of time\\
        & $\circ$ You know the paper is a big recycling program. \\ 
        & $\circ$ But they just put it in the trash \\
        & $\circ$ Or if you have a recycling bag or anything like that. \\
    \hline
    CVAE & $\circ$ Um - hum.\\
    +BOW    & $\circ$ Oh that would be nice. \\
        & $\circ$ Is that right? \\
        & $\circ$ It's a it's a luxury\\
        & $\circ$ Well that's interesting. \\
    \hline
    MTask & $\circ$ Um - hum.\\
        & $\circ$ That's right.  \\
        & $\circ$ I don't know how much money you're going to do it. \\
        & $\circ$ That's right.\\
        & $\circ$ Yeah\\
    \hline
    S2S+ & $\circ$ That's great.\\
    sampling &  $\circ$ What do you feel about this uh - huh. \\
       & $\circ$ It's just kind of a glass of town and it's even irritating to me. I don't know \\
        & $\circ$ Somebody and it's so much better. \\
        & $\circ$ It now and it's hard to go. \\
    \hline
    human & $\circ$ Yeah it is to some degree inconvenient i'd have to say that\\
    \hline
    \end{tabular}
    \caption{Example system outputs and human reference on Switchboard dataset. Controversial or offensive words are replaced by ***.}
    \label{table:example_switchboard}
\end{table}

\subsection{Human evaluation}
We randomly sampled 500 contexts from the Reddit test dataset and picked the top 1 hypothesis generated for each context ranked by $\log p(h_j|x_i) + \lambda |h_j|$. As in the automatic evaluation, we tuned $\lambda$ such that all systems have roughly the same average hypothesis length. We also randomly select one reference for each context and compare them with the systems (labeled "human" in Table~\ref{table:human})

As illustrated in Table~\ref{table:human}, the proposed model outperforms all systems except human, consistent with our automatic evaluation results. 

\begin{table}[H]
    \centering
    \small
    \begin{tabular}{p{0.16\textwidth}|p{0.07\textwidth}p{0.07\textwidth}p{0.07\textwidth}} \hline 
    
                     & relevance     & interest        & average \\ \hline
    \textsc{SpaceFusion}          & \textbf{2.72} & \textbf{2.53} & \textbf{2.63} \\ 
    CVAE+BOW         & 2.51         & 2.37          & 2.44 \\
    Multi-Task       & 2.34          & 2.14          & 2.24 \\ 
    S2S+Sampling & 2.58          & 2.43          & 2.50 \\ \hline
    human            & 3.59          & 3.41          & 3.50 \\ \hline
    
    \end{tabular}
    \caption{Performance of each model on human evaluation. The highest score, except human, in each row is in \textbf{bold}.}
    \label{table:human}
\end{table}

\section{Conclusion}
We propose  a \textsc{SpaceFusion} model to jointly optimize diversity and relevance that leverages novel regularization terms to essentially fuse the latent space of a S2S model with that of an autoencoder model. This fused latent space exhibits desirable properties such as smooth semantic interpolation between two points. The distance and direction from the predicted response vector roughly match relevance and diversity, respectively. These properties also enable intuitive visualization of the latent space. Both automatic and human evaluation results demonstrate that the proposed approach brings significant improvement compared to strong baselines in terms of both diversity and relevance. 
In future work, we will provide theoretical justification of the effectiveness of the proposed regularization terms. We expect that this technique will find application as an efficient "mixing board" for conversation that draws on multiple sources of information.

\clearpage
\newpage
\bibliography{main}
\bibliographystyle{acl_natbib}

\end{document}